\documentclass{llncs}
\usepackage[table,xcdraw]{xcolor}
\usepackage{graphicx}
\usepackage{listings}
\usepackage{bm}
\usepackage{pgfplots}
\usepackage{tikz}
\usepackage{amsmath}
\pgfplotsset{compat=newest}
\usepackage{dsfont}
\usepgfplotslibrary{groupplots}
\usepackage[utf8]{inputenc}
\usepackage{algorithm}
\usepackage{algpseudocode}
\usepackage{hyperref}
\usepackage{accents}
\usepackage{cite}
\usepackage{booktabs}
\usepackage{url}
\urldef{\mailsa}\path|aminfadaee@aut.ac.ir|
\urldef{\mailsb}\path|m.amirhaeri@utwente.nl|

\begin {document}

    \title {MULTI-LABEL CLASSIFICATION USING LINK PREDICTION}

    \author{Seyed Amin Fadaee\inst{1} \and Maryam Amir Haeri\inst{2}}

    \institute{Department of Computer Science and Information Technology, Amirkabir University of Technology, Iran; \mailsa \and
    Department of Research Methodology, Measurement and Data Analysis, University of Twente, Enschede, Netherlands; \mailsb}

    \maketitle

    \begin{abstract}
        Solving classification with graph methods has gained huge popularity in recent years.
This is due to the fact that the data can be intuitively modeled with graphs to utilize high level features to aid in solving the classification problem.
CULP which is short for Classification Using Link Prediction is a graph-based classifier.
This classifier utilizes the graph representation of the data and transforms the problem to that of link prediction where we try to find the link between an unlabeled node and the proper class node for it.
CULP proved to be highly accurate classifier and it has the power to predict the labels in near constant time.
A variant of the classification problem is multi-label classification which tackles this problem for multi-label data where an
instance can have multiple labels associated to it.
In this work, we extend the CULP algorithm to address this problem.
Our proposed extensions conveys the powers of CULP and its intuitive representation of the data in to the multi-label domain and in comparison to
some of the cutting edge multi-label classifiers, yield competitive results.
    \end {abstract}

    \section{Introduction}
    Classification is a supervised learning problem with the goal of learning a model using the labeled data.
This model can then be used to predict the labels of new data.
Binary-classification refers to a variant of this problem where a data can have one of two different values and is the most basic type of classification.
In multi-class classification, a data point can take a label from multiple choices of labels (more than two) \cite{murphy}.
\emph{Multi-label classification} is another variant of the classification problem where each data point can have multiple labels.
This problem has been utilized in applications such as labeling of multimedia resources, text categorization, genetics and biology \cite{mlc}.

There are different approaches in solving the multi-label classification problem one of which is transforming the data.
Data transformation changes the data in a way that allows a normal classifier to be used.
Two main approaches which use this technique are \emph{Binary Relevance} \cite{br} which trains a separate classifier for each label and
\emph{Label PowerSet} \cite{lp} which treats each combination of labels as a single class upon training.
The other technique is method adaptation which changes a method used for multi-class classification and makes it suitable for learning from multi-label data.
Ensemble algorithms, being transformation methods by nature, are considered as another technique in tackling this problem.
A notable example of this category is the \emph{RAkEL} \cite{rakel} algorithm.

An approach which has been researched extensively for solving multi-class classification but has not been studied for learning from multi-label data is Complex Networks.
Complex networks is a domain of data mining which deals with graphs in huge dimensionality.
When solving a machine learning task in this settings, a graph representation of the data is used to solve the problem.
These methods usually convert the original data to one or multiple graphs (with $k$NN or $r$-radius method \cite{gconversion})
and using one or more specific graph properties, they find the labels for the unlabeled data \cite{culp, schemeDataClassificationUsingRandomWalk, dataClassificationBasedOnImportance, networkBasedHighLevelDataClassification, lowHighLevelStacking, highLevelTotallyBasedOnComplexetworks}.

One of the algorithms in this domain is CULP (Classification Using Link Prediction), proposed in \cite{culp} as a high accuracy classifier for solving the problem of multi-class classification.
This algorithm utilizes complex networks, specifically the problem of link prediction to aid in solving the classification problem.
CULP models the data and labels as a graph structure called \emph{LEG} (Label Embedded Graph) which uses links between a data node and a class node
to define class membership;
this in turn converts the problem of classification to a link prediction task where the link between an unlabeled data node and a class node must be
determined.

In this work, we presented 2 extensions to the CULP algorithm, namely \emph{MiCULP} (CULP with MultI-label embedding)
and \emph{BiCULP} (CULP with BInarily dissected labels embedding) which adapt the concept of CULP to solve the
problem of multi-label classification.
These algorithms fall in to the adaptive approach of solving the multi-label classification category which was mentioned earlier.
As is elaborated in the experiments section, these extensions yield competitive results in comparison to the cutting edge algorithms for dealing with
multi-label data.

In the next section, a more detailed review on CULP, along with the complete algorithms for MiCULP and BiCULP are presented.
After that we review some of the recent works on multi-label classification and introduce some of the algorithms which will be used in experiments.
Following this will be the experiments done to evaluate the performance of the extensions on some benchmark data sets and the last sections would
conclude the paper along with setting some goals for following the works done here for the future.

    \section{Proposed Method}
    In our previous work \cite{culp}, we devised an algorithm to solve the classification problem using the
similarities between instances.
CULP (Classification Using Link Prediction) utilizes the power of complex networks -
link prediction in specific to determine the label of a data point.
From the inception of this algorithm, the extension of it for solving the problem of multi-class classification
was apparent.

CULP converts the task of classification to a link prediction problem.
This is done by utilizing the LEG data-structure (short for Label Embedded Graph).
LEG is a heterogeneous graph with two sets of edges and three sets of nodes which represent the data in form of a
graph, suitable for classification.

Given the original labeled data, $X^{(l)}$ with labels $y$ where $y_i \in \{1,2,\ldots, C\}$ with $C$ being the
number of classes along with the unlabeled data $X^{(u)}$, we can create LEG nodes and edges.

A node can belong to the training nodes set ($V_l$), testing nodes set ($V_u$) or the class nodes set ($V_c$).
Each of the instances in $X^{(l)}$ is represented by a training node and each instance in $X^{(u)}$ is
represented by
a testing node.
Considering there are $C$ labels for instances of $X^{(l)}$, $C$ class nodes would be created each of
which represents one of the labels.

The edges of the LEG comprise of similarity edges ($E_s$) and membership edges ($E_c$).
Similarity edges connect node $i$ and node $j$ (where $i,j \in \{V_l \cup V_u\}$) and a membership edge
connects a training node $i \in V_l$ to a class node $c \in V_c$;
This means that if $X^{(l)}_i$ has label $y_i=c$, there would be an edge from $i$ to $c$.

Edges in $E_s$ incorporate the similarities between instances of our data and the edges in this set are
obtained by using a graph conversion algorithm.
CULP uses an undirected version of the $k$NN graph conversion \cite{culp} but any other method is applicable as needed.
The graph conversion method uses a vector similarity function namely $s$, which can be a typical function like
Euclidean, Manhattan or Cosine similarity.

An example of creating a LEG is depicted in figure \ref{toy}.
In this example the black and white dots represent $X^{(l)}$ and the red square point is an unlabeled data.
Instances in $X^{(l)}$ has two sets of labels, white dots are labeled $1$ and the black ones have label of $2$ and
as can be seen in the graph, the corresponding training nodes are connected to their label nodes $j_1$ and $j_2$
accordingly.

\input{f-leg.tex}

Despite the heterogeneous nature of a LEG, we can define it as a simple undirected graph $G(V,E)$ where $V=V_l\cup
V_u\cup V_c$ and $E=E_s\cup E_c$.
The procedure for creating $G$ is summarized in Algorithm \ref{LEG}.

\begin{algorithm}[h!]
    \caption{constructing LEG using $X^{(l)}$, the labels $y$ and the unlabeled data $X^{(u)}$ with
    parameter $k$ and the similarity function $s$}
    \label{LEG}
    \begin{algorithmic}
        \Function{LEG}{$X^{(l)}$, $X^{(u)}$, $y$, $s$, $k$}
            \State {$X = X^{(l)}\cup X^{(u)}$
            \State $V_l \gets \{1,2,...,n\}$ \footnotesize{//Nodes are represented by numbers}}
            \State $V_u \gets \{n+1,n+2,..., n+m\}$
            \State $V_c \gets \{n+m+1,n+m+2,...,n+m+C\}$
            \State $E_c \gets \{\}$
            \For {$i\in\{1,2,...,n\}$}
                \State $E_c \gets E_c \cup(i,n+m+ y_i)$
            \EndFor
            \State $E_s \gets $kNN-CONVERT$(X,s,k)$
            \State  $V \gets V_l \cup V_u \cup V_c$
            \State  $E \gets E_s \cup E_c$
            \State \Return $G(V, E)$

        \EndFunction
    \end{algorithmic}
\end{algorithm}

After creating the LEG, we can predict the labels of the unlabeled instances by finding the most probable class
node to connect to each of the test nodes.

Using link prediction, we can easily solve this problem.
By using a local similarity measure $\lambda$ as a link predictor (e.g. Adamic-Adar index \cite{adamicAdar}), this
problem
can be solved as the following:

\begin{equation}
    \label{eq1}
    \begin{split}
        \forall X^{(u)}_i\in X^{(u)},\;\; y_i=c^* \\
        c^*= \underset{j\in V_c}{argmax}(\lambda_{i,c})\;\;
    \end{split}
\end{equation}

In the formula above, the $c^*$ is derived by finding the class node which has the highest probability of connecting
to node $i$.

Using simple link prediction schemes such as Common Neighbors \cite{commonNeighbors}, Adamic Adar Index and
Resource Allocation Index \cite{resourceAllocation}, CULP proved to be competitive against typical
classification algorithms and some more powerful graph-based classifiers.
The procedure of CULP can be seen in Algorithm \ref{culp}.

\begin{algorithm}[h]
    \caption{CULP Algorithm}
    \label{culp}
    \begin{algorithmic}
        \Function{CULP}{$X^{(l)}$, $X^{(u)}$, $y$, $s$, $k$, $\lambda$}
            \State $G \gets LEG(X^{(l)}, X^{(u)}, y, s, k)$
            \State $\hat{y}\gets \{\}$
            \For {$i \in V_u$}
                \State $c^*\gets \underset{c\in V_c}{argmax}(\lambda_{i,c})$\\
                \State $\hat{y}_i \gets c^*-(n+m)$
            \EndFor
            \State \Return $\hat{y}$
        \EndFunction
    \end{algorithmic}
\end{algorithm}

In \cite{culp}, we also extended this algorithm with CULM (CULp with Majority vote), an ensemble algorithm that
uses multiple link predictors along with a typical classifier and a majority vote scheme to drive the best
accuracy scores compare to other methods.

By extending the same principles used in CULP, we devised two algorithms for solving the problem of multi-label
classification which will be discussed in the next section.

\subsection{MiCULP Aglorithm}

In the setting of multi-label classification, given the data $X^{(l)}$, each instance in $X^{(l)}$ can have
multiple labels from the label set $L=\{1,2,\ldots,C\}$, so essentially we would have a label \emph{matrix} $Y$
instead of a label vector.
Given a training instance $X^{(l)}_i$, $Y_i$ is the label vector of $X^{(l)}_i$ where $\forall Y_{i,j}\in Y_i:
Y_{i,j}\in\{0,
1\}$.

As previously discussed, CULP derives a score for each pair of $(i,c)$ where $i\in V_u$ and $c\in V_c$.
For a multi-label data, we can tweak LEG and CULP algorithms to use a threshold on the similarities and therefore
pick multiple labels.

The extension of LEG used for this task is called \emph{MiLEG} which is short for MultI-Label Embedded Graph. The
main difference between a LEG and a MiLEG is that in a MiLEG structure, a training node $i$ can be connected to
multiple class nodes.
The procedure for creating a MiLEG is captured in Algorithm \ref{MiLEG}. \footnote{The complete code for both extension in Python can be found in \href{https://github.com/aminfadaee/mlculp}{github.com/aminfadaee/mlculp}}

\begin{algorithm}[h!]
    \caption{constructing MiLEG using $X^{(l)}$, labels $Y$, the label set $L$ and the unlabeled data $X^{
    (u)}$ with
    parameter $k$ and the similarity function $s$}
    \label{MiLEG}
    \begin{algorithmic}
        \Function{MiLEG}{$X^{(l)}$, $X^{(u)}$, $Y$, $L$, $s$, $k$}
            \State {$X = X^{(l)}\cup X^{(u)}$
            \State $V_l \gets \{1,2,...,n\}$ \footnotesize{//Nodes are represented by numbers}}
            \State $V_u \gets \{n+1,n+2,..., n+m\}$
            \State $V_c \gets \{n+m+1,n+m+2,...,n+m+C\}$
            \State $E_c \gets \{\}$
            \For {$i\in\{1,2,...,n\}$}
                \For {$y_{i,c} \in Y_i$}
                    \If {$y_{i,c} = 1$}
                        \State $E_c \gets E_c \cup(i,n+m+c)$
                    \EndIf
                \EndFor
            \EndFor
            \State $E_s \gets $kNN-CONVERT$(X,s,k)$
            \State  $V \gets V_l \cup V_u \cup V_c$
            \State  $E \gets E_s \cup E_c$
            \State \Return $G(V, E)$

        \EndFunction
    \end{algorithmic}
\end{algorithm}

As for the prediction procedure, we need to pick multiple labels that have scores beyond a defined threshold like
the following:

\begin{equation}
    \label{eq2}
    \forall X^{(u)}_i\in X^{(u)}; \forall y_{i,c}\in Y_i
    \begin{cases}
        y_{i,c}=1 & \lambda_{i,c}> t\\\\
        y_{i,c}=0 & \lambda_{i,c}\leq t
    \end{cases}
\end{equation}

In the formula above, $t$ is used as a threshold for picking labels. $t$'s optimal value can be found using a
validation scheme or have a pre-defined value depending on the data.
Using this, we can completely define the MiCULP algorithm which is depicted in Algorithm \ref{miculp}.

\begin{algorithm}[h]
    \caption{MiCULP Algorithm}
    \label{miculp}
    \begin{algorithmic}
        \Function{MiCULP}{$X^{(l)}$, $X^{(u)}$, $Y$, $L$, $s$, $k$, $\lambda$, $t$}
            \State $G \gets MiLEG(X^{(l)}, X^{(u)}, Y, L, s, k)$
            \State $\hat{Y}\gets [\; ]$
            \For {$i \in V_u$}
                \For {$c \in V_c$}
                    \State $\hat{Y}_{i,c-(n+m)}=1$ \textbf{if} $\lambda_{i,c} > t$ \textbf{else} $0$
                \EndFor
            \EndFor
            \State \Return $\hat{y}$
        \EndFunction
    \end{algorithmic}
\end{algorithm}

In this algorithm, after creating the MiLEG representation, each of the unlabeled data is assigned a label vector by comparing the score of the link between that
data node and all the label nodes against $t$.
For simplicity, we can normalize these scores to pick $t$ from $[0, 1]$ range.

\subsection{BiCULP Aglorithm}

The other extension to CULP algorithm for multi-label classification is \emph{BiCULP}.
This extension is inspired by the binary relevance method and uses the data structure \emph{BiLEG} which is short for
BInarily dissected Label Embedded Graph.

The BiLEG structure creates 2 nodes for each of the labels $c \in L$.
As stated before, each label $c$ can have a value of $0$ or $1$;
in BiLEG, each of these values are denoted by nodes $c_0$ and $c_1$.
Given this, node $i$ is connected to $c_1$ if $y_{i,c}=1$ or to $c_0$ otherwise.
The complete procedure for creating this data structure is defined in Algorithm \ref{BiLEG}.

\begin{algorithm}[h!]
    \caption{constructing BiLEG using $X^{(l)}$, the labels matrix $Y$, the label set $L$ and the unlabeled data $X^{
    (u)}$ with
    parameter $k$ and the similarity function $s$}
    \label{BiLEG}
    \begin{algorithmic}
        \Function{BiLEG}{$X^{(l)}$, $X^{(u)}$, $Y$, $L$, $s$, $k$}
            \State {$X = X^{(l)}\cup X^{(u)}$
            \State $V_l \gets \{1,2,...,n\}$ \footnotesize{//Nodes are represented by numbers}}
            \State $V_u \gets \{n+1,n+2,..., n+m\}$
            \State $V_c \gets \{n+m+1,n+m+2,...,n+m+2C\}$
            \State $E_c \gets \{\}$
            \For {$i\in\{1,2,...,n\}$}
                \For {$y_{i,c} \in Y_i$}
                    \State $E_c \gets E_c \cup(i,n+m+2c-y_{i,c})$
                \EndFor
            \EndFor
            \State $E_s \gets $kNN-CONVERT$(X,s,k)$
            \State  $V \gets V_l \cup V_u \cup V_c$
            \State  $E \gets E_s \cup E_c$
            \State \Return $G(V, E)$
        \EndFunction
    \end{algorithmic}
\end{algorithm}

In this algorithm $2C$ class nodes are present and the nodes for label $c$ are $2c$ and $2c-1$ respectively for $0$ ($c_0$) and $1$ ($c_1$) values of $c$.

Having the BiLEG, the BiCULP algorithm easily determines the labels of the node $i$ in the following way:

\begin{equation}
    \label{eq3}
    \forall X^{(u)}_i\in X^{(u)}; \forall y_{i,c}\in Y_i
    \begin{cases}
        y_{i,c}=1 & \lambda_{i,c_1}> \lambda_{i,c_0}\\\\
        y_{i,c}=0 & otherwise
    \end{cases}
\end{equation}

An apparent gain of using this approach compared to that of MiCULP is that finding a suitable threshold is no longer needed.
Furthermore, this approach utilizes the lack of labels in prediction as well which can be beneficial in presence of imbalanced classes.
The complete algorithm for BiCULP is presented in Algorithm \ref{biculp}.

\begin{algorithm}[h]
    \caption{BiCULP Algorithm}
    \label{biculp}
    \begin{algorithmic}
        \Function{BiCULP}{$X^{(l)}$, $X^{(u)}$, $Y$, $L$, $s$, $k$, $\lambda$}
            \State $G \gets BiLEG(X^{(l)}, X^{(u)}, Y, L, s, k)$
            \State $\hat{Y}\gets [\; ]$
            \For {$i \in V_u$}
                \For {$c-1,c \in V_c$}
                        \State $\hat{Y}_{i,\frac{c}{2}-(n+m)}=1$ \textbf{if} $\lambda_{i,c-1} > \lambda_{i,c}$ \textbf{else} $0$
                \EndFor
            \EndFor
            \State \Return $\hat{y}$
        \EndFunction
    \end{algorithmic}
\end{algorithm}

\subsection{Toy Example}

In this section we introduce a toy example to elaborate the concepts discussed in MiCULP and BiCULP extensions.
As can be seen in table \ref{toy-data}, there are 6 points having at least one of three labels of $a$, $b$ and
$c$ and a seventh point (denoted by $i$ from here on), which the labels for is unknown.

\begin{table*}[t]
    \small
    \caption{A simple set of multi-label data.}\label{toy-data}
    \centering
\begin{tabular}{c|ccccccc}
    \textbf{$x_1$} & 1 & 0 & 1 & 1  & 0 & 0  & 0  \\
    \textbf{$x_2$} & 0 & 1 & 2 & -2 & 3 & -3 & -1 \\
    \midrule
    \textbf{$a$}   & 1 & 1 & 1 & 0  & 1 & 0  & ?? \\
    \textbf{$b$}   & 1 & 1 & 1 & 1  & 0 & 0  & ?? \\
    \textbf{$c$}   & 1 & 0 & 0 & 1  & 0 & 1  & ?? \\
\end{tabular}

\end{table*}

\begin{figure}[h]
    \centering

    \newcommand{\scale}{1}
    \newcommand{\bscale}{1}
    \definecolor{ubuntu}{rgb}{0.2,0.2,0.2}
    \definecolor{light}{rgb}{0.6,0.6,0.6}
    \begin{tikzpicture}
        \begin{groupplot}[group style={group size=2 by 1},height=7cm, width=\columnwidth/1.25]
            \nextgroupplot[
            hide x axis,
            hide y axis,
            title={$A$},
            xmin=-7, xmax=7,
            ymin=-5, ymax=5,
            ]
            \draw[black] (axis cs:4,2) --(axis cs:1,0);
            \draw[black] (axis cs:4,2) --(axis cs:0,1);
            \draw[black] (axis cs:4,2) --(axis cs:1,2);
            \draw[black] (axis cs:4,2) --(axis cs:0,3);

            \draw[black] (axis cs:4,0) --(axis cs:1,0);
            \draw[black] (axis cs:4,0) --(axis cs:0,1);
            \draw[black] (axis cs:4,0) --(axis cs:1,2);
            \draw[black] (axis cs:4,0) --(axis cs:1,-2);

            \draw[black] (axis cs:4,-2) --(axis cs:1,0);
            \draw[black] (axis cs:4,-2) --(axis cs:1,-2);
            \draw[black] (axis cs:4,-2) --(axis cs:0,-3);

            \draw[black] (axis cs:0,3) --(axis cs:1,2);
            \draw[black] (axis cs:0,3) --(axis cs:0,1);
            \draw[black] (axis cs:0,1) --(axis cs:1,0);
            \draw[black] (axis cs:0,1) --(axis cs:0,-1);
            \draw[black] (axis cs:0,-1) --(axis cs:1,-2);
            \draw[black] (axis cs:0,-1) --(axis cs:0,-3);
            \draw[black] (axis cs:1,2) --(axis cs:0,1);
            \draw[black] (axis cs:1,2) --(axis cs:1,0);
            \draw[black] (axis cs:1,-2) --(axis cs:0,-3);
            \draw[black] (axis cs:1,-2) --(axis cs:1,0);
            \draw[black] (axis cs:0,-1) --(axis cs:1,0);

            \path [draw=black] (axis cs:5,8.66025403784439)
            --(axis cs:7.07106781186547,7.07106781186548);

            \node at (axis cs:1,0)[fill=ubuntu, circle, scale=\scale, minimum size=0.2cm, text=black, draw=black]{};
            \node at (axis cs:0,1)[fill=ubuntu, circle, scale=\scale, minimum size=0.2cm, text=black, draw=black]{};
            \node at (axis cs:0,-1)[fill=red, circle, scale=\scale, minimum size=0.1cm, text=white, draw=white]{$i$};
            \node at (axis cs:1,2)[fill=ubuntu, circle, scale=\scale, minimum size=0.2cm, text=black, draw=black]{};
            \node at (axis cs:1,-2)[fill=ubuntu, circle, scale=\scale, minimum size=0.2cm, text=black, draw=black]{};
            \node at (axis cs:0,3)[fill=ubuntu, circle, scale=\scale, minimum size=0.2cm, text=black, draw=black]{};
            \node at (axis cs:0,-3)[fill=ubuntu, circle, scale=\scale, minimum size=0.2cm, text=black, draw=black]{};

            \node at (axis cs:4,2)[fill=white, circle, scale=\bscale, minimum size=0.6cm, text=black, draw=black]{$a$ };
            \node at (axis cs:4,0)[fill=white, circle, scale=\bscale, minimum size=0.5cm, text=black, draw=black]{$b$ };
            \node at (axis cs:4,-2)[fill=white, circle, scale=\bscale, minimum size=0.6cm, text=black, draw=black]{$c$ };

            \nextgroupplot[
            hide x axis,
            hide y axis,
            title={$B$},
            xmin=-7, xmax=7,
            ymin=-5, ymax=5,
            ]
            \draw[black] (axis cs:0,3) --(axis cs:-4,2);
            \draw[black] (axis cs:1,2) --(axis cs:-4,2);
            \draw[black] (axis cs:0,1) --(axis cs:-4,2);

            \draw[black] (axis cs:1,-2) --(axis cs:-4,-2);
            \draw[black] (axis cs:0,-3) --(axis cs:-4,-2);

            \draw[black] (axis cs:0,-3) --(axis cs:-4,0);
            \draw[black] (axis cs:0,3) --(axis cs:-4,0);

            \draw[black] (axis cs:4,2) --(axis cs:1,0);
            \draw[black] (axis cs:4,2) --(axis cs:0,1);
            \draw[black] (axis cs:4,2) --(axis cs:1,2);
            \draw[black] (axis cs:4,2) --(axis cs:0,3);

            \draw[black] (axis cs:4,0) --(axis cs:1,0);
            \draw[black] (axis cs:4,0) --(axis cs:0,1);
            \draw[black] (axis cs:4,0) --(axis cs:1,2);
            \draw[black] (axis cs:4,0) --(axis cs:1,-2);

            \draw[black] (axis cs:4,-2) --(axis cs:1,0);
            \draw[black] (axis cs:4,-2) --(axis cs:1,-2);
            \draw[black] (axis cs:4,-2) --(axis cs:0,-3);

            \draw[black] (axis cs:0,3) --(axis cs:1,2);
            \draw[black] (axis cs:0,3) --(axis cs:0,1);
            \draw[black] (axis cs:0,1) --(axis cs:1,0);
            \draw[black] (axis cs:0,1) --(axis cs:0,-1);
            \draw[black] (axis cs:0,-1) --(axis cs:1,-2);
            \draw[black] (axis cs:0,-1) --(axis cs:0,-3);
            \draw[black] (axis cs:1,2) --(axis cs:0,1);
            \draw[black] (axis cs:1,2) --(axis cs:1,0);
            \draw[black] (axis cs:1,-2) --(axis cs:0,-3);
            \draw[black] (axis cs:1,-2) --(axis cs:1,0);
            \draw[black] (axis cs:0,-1) --(axis cs:1,0);

            \path [draw=black] (axis cs:5,8.66025403784439)
            --(axis cs:7.07106781186547,7.07106781186548);

            \node at (axis cs:1,0)[fill=ubuntu, circle, scale=\scale, minimum size=0.2cm, text=black, draw=black]{};
            \node at (axis cs:0,1)[fill=ubuntu, circle, scale=\scale, minimum size=0.2cm, text=black, draw=black]{};
            \node at (axis cs:0,-1)[fill=red, circle, scale=\scale, minimum size=0.1cm, text=white, draw=white]{$i$};
            \node at (axis cs:1,2)[fill=ubuntu, circle, scale=\scale, minimum size=0.2cm, text=black, draw=black]{};
            \node at (axis cs:1,-2)[fill=ubuntu, circle, scale=\scale, minimum size=0.2cm, text=black, draw=black]{};
            \node at (axis cs:0,3)[fill=ubuntu, circle, scale=\scale, minimum size=0.2cm, text=black, draw=black]{};
            \node at (axis cs:0,-3)[fill=ubuntu, circle, scale=\scale, minimum size=0.2cm, text=black, draw=black]{};

            \node at (axis cs:4,2)[fill=white, circle, scale=\bscale, minimum size=0.6cm, text=black, draw=black]{$a_1$ };
            \node at (axis cs:4,0)[fill=white, circle, scale=\bscale, minimum size=0.5cm, text=black, draw=black]{$b_1$ };
            \node at (axis cs:4,-2)[fill=white, circle, scale=\bscale, minimum size=0.6cm, text=black, draw=black]{$c_1$ };

            \node at (axis cs:-4,2)[fill=white, circle, scale=\bscale, minimum size=0.6cm, text=black, draw=black]{$c_0$ };
            \node at (axis cs:-4,0)[fill=white, circle, scale=\bscale, minimum size=0.5cm, text=black, draw=black]{$b_0$ };
            \node at (axis cs:-4,-2)[fill=white, circle, scale=\bscale, minimum size=0.6cm, text=black, draw=black]{$a_0$ };

        \end{groupplot}

    \end{tikzpicture}

    \caption{Toy examples of LEG extensions. A- The MiLEG representation B- The BiLEG representation} \label{toy}
\end{figure}
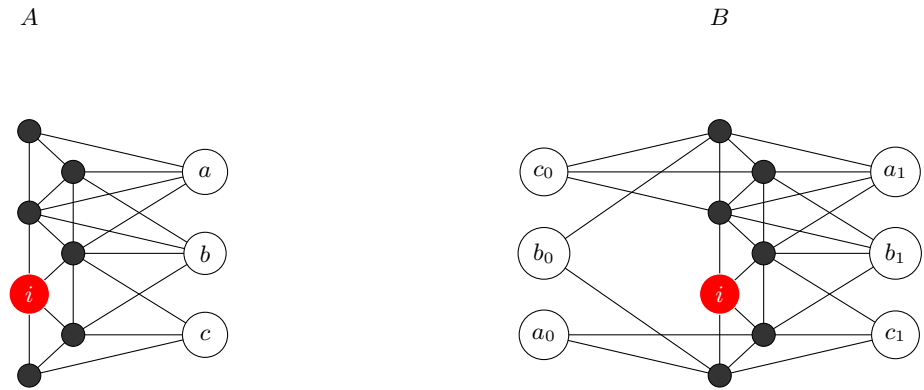

The first step in prediction of $i$'s labels with the proposed algorithms is the creation of MiLEG and BiLEG
structures which are shown in figure \ref{toy}.
The left graph which depicts MiCULP has 3 class nodes and that of the BiLEG in the right has 3 pairs of class nodes.
In both graphs, each training point is connected to its corresponding label nodes and non-class nodes are connected to each other
via the similarity links.

Given a common neighbors link prediction strategy, we can find the scores of $(i,a)$, $(i,b)$ and $(i,c)$ in MiLEG as $2$, $3$ and $3$ respectively.
With a $t$ with value of $3$, we can find the labels of $i$ as ${b,c}$ which matches our expectations of this node.

Using the same strategy in BiLEG results in $\lambda_{i,a_0}=2$, $\lambda_{i,a_1}=2$, $\lambda_{i,b_0}=1$,  $\lambda_{i,b_1}=3$,
$\lambda_{i,c_0}=1$, $\lambda_{i,c_1}=3$.
Comparing each pair of values will yield the same results of ${b,c}$ as the labels for $i$.

    \section{Related Work}
    MiCULP and BiCULP algorithms are extended versions of CULP algorithm \cite{culp} for multi-label classification.
CULP proved to be a highly accurate classifier, able to utilize the high-level features in data and fast in prediction of unlabeled data.
The inner logic of CULP makes it intuitively extendable for multi-label classification which leads to capturing its powers in solving this problem.
In the rest of this section, some of the state of the art algorithms in multi-label classification are reviewed to provide some alternatives to compare
against our work.

\emph{ACkEL} \cite{ackel} is an improvement on the RAkEL algorithm which trains different classifiers on subsets of the data using the
label powerset method.
The ACkEL algorithm uses kernel support vector machines as the classifiers and uses the
Fisher's linear discriminant ratio to determine the \emph{separability} of the classes and joint entropy to describe the
\emph{imbalance level} in the data.
These two parameters are linearly combined to evaluate the quality of a subset of the data.
Furthermore, this algorithm uses active learning scheme to select samples based on the discriminant ratio and the joint entropy.
In the experiments, we used the version of ACkEL which uses overlapping subsets of data, namely \emph{ACkELo} because it outperforms
the other variant - \emph{ACkELd}.

In \cite{boomer}, a framework for optimizing different multi-label learning loss functions by finding rule sets to optimize it with gradient boosting,
along with \emph{BOOMER}, the instantiation of this framework is proposed.
This algorithm is able to minimize both decomposable and non-decomposable loss functions and is another algorithm which we used in our experiments.

\emph{MLWSE} \cite{mlwse} is a stacked ensemble algorithm which uses pairwise label correlations and it uses label-specific
meta features to facilitate the classification process.
In this algorithm, two classifiers with highly correlated labels share high similar weights as opposed to two classifiers with weakly correlated labels.
In our work we chose the \emph{MLWSE-L2} variant of this algorithm to compare against which uses $L2$ sparsity regularization.
This regularization is used to prevent the stacked ensemble from combining all the base classifiers.

In \cite{entropycc}, an ordering scheme for classifier chains is proposed.
This approach finds a single order for the chain based on conditional entropy of the class labels and does not train any
additional classifier to aid in this process.
The authors showed that their method can outperform many ordering schemes in the literature.

    \section{Experimental Results}
    In order to evaluate the MiCULP and BiCULP algorithms on a practical level, in this section we apply them on some benchmark data sets.
The data sets used in our experiments are \emph{birds} \cite{birds}, \emph{emotions} \cite{emotions}, \emph{enron} \cite{enron}, \emph{flags} \cite{flags} and
\emph{scene} \cite{scene}.
These data sets along with some of their properties are listed in Figure \ref{datasets}.

\begin{table*}[ht]
    \small
    \caption{Dataset used for experiments. $n$ is the number of samples, $f$ is the number of features, $C$ is the number of labels and $|L|$ is the total number of combinations of labels.}\label{datasets}
    \centering
\begin{tabular}{p{3cm}lllr}
    \toprule
    \textbf{Dataset} & \textbf{n} & \textbf{f} & \textbf{C} & \textbf{|L|} \\
    \midrule
    Birds            & 645        & 260        & 19         & 133          \\
    \midrule
    Emotions         & 593        & 72         & 6          & 27           \\
    \midrule
    Enron            & 1702       & 1001       & 53         & 15806        \\
    \midrule
    Flags            & 194        & 19         & 7          & 54           \\
    \midrule
    Scene            & 2407       & 294        & 6          & 15           \\
    \bottomrule
\end{tabular}
\end{table*}

As mentioned previously, the algorithms used for comparing our results with are ACkELo \cite{ackel},
MLWSE-L2 \cite{mlwse} and BOOMER \cite{boomer}.
As for the evaluation criteria, we used hamming loss, example based F1 score (F1 for short), micro-averaging F1 score
(micro-F1) and macro-averaging F1 (macro-F1).
The results of each of these metrics are presented in Tables~\ref{result-hamming},~\ref{result-f1},~\ref{result-micro-f1}
and~\ref{result-macro-f1}.

\begin{table*}[ht]
    \scriptsize
    \caption{Hamming Results}\label{result-hamming}
    \centering
\begin{tabular}{lccccc}
    \toprule
    \textbf{Dataset} & \textbf{MiCULP}          & \textbf{BiCULP}          & \textbf{ACkELo}          & \textbf{MLWSE-L2}        & \textbf{BOOMER}          \\
    \midrule
	Birds            & $0.050 \pm 0.007_5$      & $0.047 \pm 0.007_4$      & \bm{$0.040 \pm 0.001_1$} & $0.045 \pm 0.001_3$      & $0.041 \pm 0.007_2$      \\
    \midrule
    Emotions         & $0.188 \pm 0.029_2$      & \bm{$0.175 \pm 0.019_1$} & $0.199 \pm 0.008_5$      & $0.193 \pm 0.007_4$      & $0.189 \pm 0.020_3$      \\
    \midrule
	Enron            & $0.051 \pm 0.002_4$      & $0.048 \pm 0.003_3$      & $0.057 \pm 0.000_5$      & \bm{$0.046 \pm 0.000_1$} & $0.047 \pm 0.002_2$      \\
    \midrule
    Flags            & $0.288 \pm 0.026_5$      & $0.264 \pm 0.028_3$      & $0.268 \pm 0.012_4$      & $0.257 \pm 0.014_2$      & \bm{$0.240 \pm 0.040_1$} \\
    \midrule
    Scene            & \bm{$0.070 \pm 0.006_1$} & $0.073 \pm 0.008_2$      & $0.079 \pm 0.001_3$      & $0.085 \pm 0.003_5$      & $0.081 \pm 0.006_4$      \\
    \midrule
    \textbf{Rank}    & $3.4$                    & $2.6$                    & $3.6$                    & $3.0$                    & $2.4$                    \\
    \bottomrule
\end{tabular}

\end{table*}

\begin{table*}[ht]
    \scriptsize
    \caption{F1 Results}\label{result-f1}
    \centering
\begin{tabular}{lccccc}
    \toprule
    \textbf{Dataset} & \textbf{MiCULP}          & \textbf{BiCULP}     & \textbf{ACkELo}          & \textbf{MLWSE-L2}   & \textbf{BOOMER}          \\
    \midrule
	Birds            & $0.547 \pm 0.045_3$      & $0.608 \pm 0.040_2$ & $0.149 \pm 0.006_4$      & $0.140 \pm 0.009_5$ & \bm{$0.619 \pm 0.058_1$} \\
    \midrule
    Emotions         & $0.683 \pm 0.033_2$      & $0.654 \pm 0.040_3$ & \bm{$0.718 \pm 0.008_1$} & $0.614 \pm 0.014_4$ & $0.607 \pm 0.044_5$      \\
    \midrule
    Enron            & \bm{$0.579 \pm 0.022_1$} & $0.503 \pm 0.031_3$ & $0.437 \pm 0.014_5$      & $0.576 \pm 0.006_2$ & $0.493 \pm 0.024_4$      \\
    \midrule
    Flags            & $0.728 \pm 0.030_2$      & $0.712 \pm 0.036_5$ & \bm{$0.736 \pm 0.013_1$} & $0.721 \pm 0.025_3$ & $0.720 \pm 0.051_4$      \\
    \midrule
    Scene            & \bm{$0.805 \pm 0.014_1$} & $0.738 \pm 0.020_3$ & $0.788 \pm 0.003_2$      & $0.672 \pm 0.010_4$ & $0.652 \pm 0.027_5$      \\
    \midrule
    \textbf{Rank}    & $1.8$                    & $3.2$               & $2.6$                    & $3.6$               & $3.8$                    \\
    \bottomrule
\end{tabular}

\end{table*}

\begin{table*}[hb!]
    \scriptsize
    \caption{Micro F1 Results}\label{result-micro-f1}
    \centering
\begin{tabular}{lccccc}
    \toprule
    \textbf{Dataset} & \textbf{MiCULP}          & \textbf{BiCULP}          & \textbf{ACkELo}          & \textbf{MLWSE-L2}   & \textbf{BOOMER}          \\
    \midrule
    Birds            & $0.418 \pm 0.070_3$      & \bm{$0.437 \pm 0.076_1$} & $0.408 \pm 0.011_4$      & $0.359 \pm 0.027_5$ & $0.421 \pm 0.069_2$      \\
    \midrule
    Emotions         & $0.704 \pm 0.025_2$      & $0.698 \pm 0.037_3$      & \bm{$0.715 \pm 0.005_1$} & $0.658 \pm 0.013_5$ & $0.667 \pm 0.040_4$      \\
    \midrule
    Enron            & \bm{$0.576 \pm 0.020_1$} & $0.521 \pm 0.029_4$      & $0.443 \pm 0.017_5$      & $0.566 \pm 0.004_2$ & $0.537 \pm 0.020_3$      \\
    \midrule
    Flags            & $0.740 \pm 0.031_2$      & $0.731 \pm 0.038_5$      & $0.733 \pm 0.012_4$      & $0.737 \pm 0.017_3$ & \bm{$0.751 \pm 0.046_1$} \\
    \midrule
    Scene            & \bm{$0.794 \pm 0.017_1$} & $0.774 \pm 0.027_3$      & $0.777 \pm 0.002_2$      & $0.733 \pm 0.009_5$ & $0.739 \pm 0.023_4$      \\
    \midrule
    \textbf{Rank}    & $1.8$                    & $3.2$                    & $3.2$                    & $4.0$               & $2.8$                    \\
    \bottomrule
\end{tabular}

\end{table*}

\begin{table*}[hb!]
    \scriptsize
    \caption{Macro F1 Results}\label{result-macro-f1}
    \centering
\begin{tabular}{lccccc}
    \toprule
    \textbf{Dataset} & \textbf{MiCULP}          & \textbf{BiCULP}          & \textbf{ACkELo}          & \textbf{MLWSE-L2}        & \textbf{BOOMER}     \\
    \midrule
    Birds            & $0.351 \pm 0.073_2$      & \bm{$0.352 \pm 0.075_1$} & $0.283 \pm 0.013_4$      & $0.133 \pm 0.010_5$      & $0.331 \pm 0.084_3$ \\
    \midrule
    Emotions         & $0.688 \pm 0.027_2$      & $0.667 \pm 0.041_3$      & \bm{$0.701 \pm 0.007_1$} & $0.584 \pm 0.013_5$      & $0.641 \pm 0.042_4$ \\
    \midrule
    Enron            & $0.298 \pm 0.032_3$      & $0.301 \pm 0.023_2$      & $0.115 \pm 0.004_5$      & \bm{$0.547 \pm 0.005_1$} & $0.272 \pm 0.042_4$ \\
    \midrule
    Flags            & $0.630 \pm 0.066_4$      & $0.605 \pm 0.080_5$      & $0.640 \pm 0.021_3$      & \bm{$0.711 \pm 0.025_1$} & $0.648 \pm 0.067_2$ \\
    \midrule
    Scene            & \bm{$0.800 \pm 0.018_1$} & $0.782 \pm 0.023_3$      & $0.782 \pm 0.002_2$      & $0.665 \pm 0.010_5$      & $0.742 \pm 0.022_4$ \\
    \midrule
    \textbf{Rank}    & $2.6$                    & $2.8$                    & $3.0$                    & $3.4$                    & $3.4$               \\
    \bottomrule
\end{tabular}

\end{table*}

The parameters used in MiCULP and BiCULP algorithms are $k$ ($1\leq k \leq 45$), $\lambda$ which is one of
Common Neighbors, Resource Allocation, Adamic Adar or Compatibility Score \cite{culp}, the vector similarity function which can be
Cosine, Manhattan or Euclidean.
For each data set, the parameters are tuned via a 5-Fold Cross Validation.
After finding the parameters, 30 runs of 10-Fold cross validation which amounts to 300 total runs are done.

For the competitions, if the metric for a data set is presented in their work, that score is presented as it is.
For other cases the default parameters of each algorithm is used.

As can be seen from the rankings, in hamming loss, MiCULP has the fourth best loss.
In F1, micro F1 and macro F1 metrics, MiCULP out performs all other algorithms.
Also for the scene data set, MiCULP is out performing other methods.
Another thing that can be noticed from the results is that although BiCULP could not outperform MiCULP, it holds the second best score for hamming loss and macro F1.

    \section{Conclusion and Future Works}
    In this work we extended the CULP classifier to solve the problem of multi-label classification.
Our method uses graph representation of the data and similarities between instances as high-level features to classify a multi-label instance effectively.
As complex data needs complex features, the graph representation can capture these features intuitively and use it as an advantage to aid in the classification.
We proposed MiCULP and BiCULP algorithms which differ in the modeling of their graph representation and our experiments proved the effectiveness of these algorithms, specially
MiCULP in dealing with multi-label data.

We are planning to extend the works done here in multiple ways, firstly we can use multiple link predictor to classify the
test nodes (similar to \emph{CULM}, another extension of CULP \cite{culp}).

Another idea is to extend the BiCULP algorithms to solve the \emph{Multi-Dimension Classification} problem, a variant of multi-label classification
where a label has more than 2 values.
As BiCULP creates a node for each value, creating nodes for more than zero and one values can be easily achieved.

Finally by defining a scheme to connect the class nodes together, we can incorporate the label correlation of the data in the MiLEG or BiLEG data structures.
This can lead to an even higher accuracy and a model which defines the inner relations of the labels.

    \bibliography{./References}
    \bibliographystyle{ieeetr}
\end{document}